\documentclass[conference, a4paper]{IEEEtran}
\IEEEoverridecommandlockouts

\usepackage{cite}
\usepackage{hyperref}
\usepackage{amsmath,amssymb,amsfonts}
\usepackage{algorithmic}
\usepackage{algorithm}
\usepackage{graphicx}
\usepackage[numbers,sort&compress]{natbib}
\usepackage{textcomp}
\usepackage[section]{placeins}
\usepackage{xcolor}
\usepackage{booktabs}
\usepackage{changes}

\usepackage[left=1.62cm,right=1.62cm,top=1.9cm,bottom=5cm]{geometry}
\graphicspath{ {./img/} }
\def\BibTeX{{\rm B\kern-.05em{\sc i\kern-.025em b}\kern-.08em
    T\kern-.1667em\lower.7ex\hbox{E}\kern-.125emX}}

\DeclareRobustCommand*{\IEEEauthorrefmark}[1]{%
\raisebox{0pt}[0pt][0pt]{\textsuperscript{\footnotesize\ensuremath{#1}}}}

\begin{document}
%
\title{STDA-Meta: A Meta-Learning Framework for Few-Shot Traffic Prediction}

%
%

\author{
\IEEEauthorblockN{Maoxiang Sun\IEEEauthorrefmark{1},
Weilong Ding\IEEEauthorrefmark{1}\thanks{Corresponding author: dingweilong@ncut.edu.cn},
Tianpu Zhang\IEEEauthorrefmark{1},
Zijian Liu\IEEEauthorrefmark{1},
Mengda Xing\IEEEauthorrefmark{1}
}
\IEEEauthorblockA{\IEEEauthorrefmark{1}School of Information Science and Technology, North China University of Technology, Beijing, China}
}

\maketitle              
\begin{abstract}
As the development of cities, traffic congestion becomes an increasingly pressing issue, and traffic prediction is a classic method to relieve that issue. Traffic prediction is one specific application of spatio-temporal prediction learning, like taxi scheduling, weather prediction, and ship trajectory prediction. Against these problems, classical spatio-temporal prediction learning methods including deep learning, require large amounts of training data. In reality, some newly developed cities with insufficient sensors would not hold that assumption, and the data scarcity makes predictive performance worse. In such situation, the learning method on insufficient data is known as few-shot learning (FSL), and the FSL of traffic prediction remains challenges. On the one hand,  graph structures' irregularity and dynamic nature of  graphs cannot hold the performance of spatio-temporal learning method. On the other hand, conventional domain adaptation methods cannot work well on insufficient training data, when transferring knowledge from different domains to the intended target domain.To address these challenges, we propose a novel spatio-temporal domain adaptation (STDA) method that learns transferable spatio-temporal meta-knowledge from data-sufficient cities in an adversarial manner. This learned meta-knowledge can improve the prediction performance of data-scarce cities. Specifically, we train the STDA model using a Model-Agnostic Meta-Learning (MAML) based episode learning process, which is a model-agnostic meta-learning framework that enables the model to solve new learning tasks using only a small number of training samples. We conduct numerous experiments on four traffic prediction datasets, and our results show that the prediction performance of our model has improved by 7\% compared to baseline models on the two metrics of MAE and RMSE.

\end{abstract}

%
\section{Introduction}
Applications such as traffic prediction \cite{b1,b2}, traffic scheduling \cite{b3,b4}, congestion prediction \cite{b5}, and automated vehicles \cite{b36} have been studied on massive data using machine learning methods, all falling under the umbrella of Intelligent Transportation Systems (ITS) \cite{b32, b33}. In scenarios with abundant data, the mentioned applications, primarily based on traditional machine learning approaches, exhibit robust performance. The challenge arises when data is insufficient, as existing machine learning algorithms usually require a significant amount of data to train models effectively. The reliance on large volumes of data can be limiting, especially in real-world scenarios where data availability is insufficient. This constraint underscores the need for exploring more efficient and effective learning approaches that can perform well even in data-scarce conditions. The reasons for the poor performance of traditional machine learning in scenarios with insufficient data are twofold. Firstly, they often fail to capture the spatio-temporal correlations between different cities, making it difficult to achieve better results from prior knowledge. Secondly, a small amount of data makes it challenging to achieve thorough gradient descent, leading to the phenomenon of overfitting.

Recently, there has been some related work on how to improve the efficiency of spatio-temporal graph learning. Yu \cite{b31} proposed a proposed a general traffic prediction method, which can extract spatio-temporal feature efficiently, but it has no good effect in the case of few-shot learning. Yao \cite{b37} first used auxiliary graphs for few-shot learning, but due to the limited parameters of the model, they could only use the model for classification problems and could only get the transfer knowledge from one source graph. A summary of the existing research indicates that the majority is centered around node classification problems. In our case, we are tackling a node regression problem for traffic prediction. The current models show deficiencies in terms of low robustness and accuracy, rendering them less appropriate for traffic prediction scenarios.

In this paper, we address the challenge of few-shot learning on traffic graph by transferring meta knowledge between different cities. Two main challenges need to be overcome: (i) Capturing the spatio-temporal domain-invariant correlations between the source and target cities. This requires an efficient method to capture the spatial and temporal characteristics of different cities at each time step, without incurring expensive computation costs. (ii) Training a model’s parameters in a way that allows for fast adaptation to new tasks with only a small number of gradient updates. Overcoming these challenges is crucial to achieving improved few-shot learning performance on traffic graph.

In response to the above challenges, we propose an effective and novel framework called \textbf{S}patio-\textbf{T}emporal \textbf{D}omain \textbf{A}daptation \textbf{Meta}-Learning(\textbf{STDA-Meta}) framework, which consists of a spatial-Temporal adversarial adaptation module(\textbf{STDA}) and a Meta-Learning framework(\textbf{Meta}). Specifically, STDA is to solve the first challenge. Inspired by GANs \cite{b9}, it distinguishes the spatial-temporal features from the source and target cities at each time step via adversarial classification to capture spatial-temporal transferable features in an efficient manner. Meanwhile, Model-Agnostic Meta-Learning(MAML) \cite{b10} framework employed is to optimize model's updates, in which a small number of gradient updates will lead to fast learning on a new task.

In summary, the main contributions of our work are as follows:
\begin{itemize}
\item We apply domain adaptation in spatio-temporal few-shot prediction, focusing on urban traffic data. Our method fully exploits the inherent structural similarities found within the graph representations of different urban traffic datasets. Generative Adversarial Networks (GANs) based data augmentation for graph few-shot learning achieves significantly improvement of predictive results, which has been rigorously substantiated through extensive ablation experiments. 
\item We merge Model-Agnostic Meta-Learning (MAML) into domain adaptation, and prove its substantial advantages in few-shot learning. By integrating MAML, rapid adaptation to new tasks on minimal data can be completed, significantly improving domain adaptation's efficiency.
Such promising solution to address few-shot challenges across various domains has been validated through extensive experiment on real-world datasets from different cities.
\item Our proposed method shows effective on traffic speed datasets from different cities, and proves  accurate to forecast traffic speed of cities with limited available data. Experimental results show that our model outperforms baseline models by 7\% in terms of prediction accuracy.
\end{itemize}


\section{RELATED WORK}
In this section, we review relevant work from two aspects. The first is spatio-temporal graph learning, which has been widely adopted in traffic prediction. 
However, when data is scarce, their performance degrade due to overfitting. To address this issue, graph few-shot learning, as the second type discussed here, developed currently. 
Through the prior knowledge of graph structure and the similarity between different graphs, such methods have shown potential advantages. 

\subsection{Spatio-Temporal Graph Learning}
Before deep-learning based approaches became popular, statistical methods were widely used in the field for spatio-temporal graph learning. Classical statistical methods such as ARIMA (Autoregressive Integrated Moving Average model) \cite{b6,b7,b8}, VAR (Vector autoregressive models) \cite{b11}, and HA (Historical Average) \cite{b12} were used to predict traffic. The advantage of statistical methods is that they depend on a single factor, are easily implemented, and quickly compute results. However, with the increasing popularity of deep learning, researchers have achieved better prediction results by using the ability of LSTM (long short-term memory) \cite{b13} and GRU (Gated Recurrent Unit) \cite{b14} to model complex functions and capture the dynamic time relationships. To capture spatial features, researchers divide the highway network into grids and use CNN \cite{b15} to extract the spatial relationship between adjacent toll stations for prediction results. However, CNN maps the traffic flow prediction problem in a non-Euclidean space to a Euclidean space, which results in a loss of spatial information. More recently, with the popularity of GCN \cite{b16}, some works such as \cite{b17,b18} combine GCN and LSTM to improve the performance of traffic prediction.

In brief, deep learning based spatio-temporal graph learning methods rely on a large amount of training data, and struggle to generalize to scenarios with insufficient data, such as traffic prediction problems.

\subsection{Graph Few-Shot Learning}

Few-shot learning has been widely adopted in various fields, such as natural language processing, computer vision, and robotics. It is an effective approach to learning from insufficient labeled data, which is often encountered in real-world applications. Among the existing few-shot learning methods, meta-learning-based models have shown promising results in addressing the problem of few-shot learning.

Few studies have explored the application of few-shot learning in the field of traffic prediction, especially in the context of spatio-temporal graph learning. Recently, there has been some progress in this direction. For example, the Meta-GNN model \cite{b21} and RALE \cite{b22} have been proposed to handle few-shot learning problems in the context of graph neural networks. These models use meta-learning techniques to learn to adapt to new tasks with few labeled examples.

Furthermore, GDN \cite{b23} is another few-shot learning model that can be used for anomaly detection on traffic graphs. This model can utilize knowledge from auxiliary networks to improve the robustness of the anomaly detection task.

Despite these promising results, the application of graph few-shot learning to traffic domain have been limited in scope and may not be sufficiently robust for real-world traffic problems. Most of them mainly focus on node classification problem rather than node regression problem in traffic prediction. 
Therefore, more effort is needed for graph few-shot learning of traffic prediction to handle node regression tasks in real-world scenarios.

\section{Problem Formulation}
\subsection{Traffic Prediction on Road Graphs}
Traffic prediction is a fundamental problem in spatio-temporal prediction, where the aim is to predict traffic status such as traffic speed, traffic flow, traffic demand, and travle time. Historical data from the previous \textit{H} time steps is utilized to predict future traffic status for the next \textit{M} time steps. 
It can be typically expressed as the maximum log-probability of the future traffic status on given the historical traffic data as the following formula. Here, $v_t \in \mathbb{R}^{n}$ represents the observed traffic status for the whole traffic network at time step $t$, where $n$ is the total number of observation nodes in the network.

\begin{equation}
\begin{array}{l}
\hat{v}_{t+1}, \ldots, \hat{v}_{t+M}= \\
\underset{v_{t+1}, \ldots, v_{t+M}}{\arg \max } \log P\left(v_{t+1}, \ldots, v_{t+M} \mid v_{t-H+1}, \ldots, v_{t}\right),
\end{array}
\end{equation}

In this paper, we focus on constructing a traffic graph based on geographic relations. The resulting graph is undirected, reflecting the fact that any two points in the traffic network are potentially connected to each other. Specifically, at time step $t$, the traffic graph is represented by an undirected graph $\mathcal{G}_t = (\mathcal{V}_t, \mathcal{E}, W)$, where $\mathcal{V}_t$ represents a collection of all the sensors in the traffic graph, which correspond to observable nodes in the graph. $\mathcal{E}$ is a collection of all the edges in the traffic network, and $W \in \mathbb{R}^{n \times n}$ denotes the weighted adjacency matrix of $\mathcal{G}_t$.

\subsection{Spatio-Temporal Graph Few-Shot Learning}
Cities are identified as two types in this study based on the amount of traffic data available. A source city is one that has a large amount of traffic data, while a target city has a relatively small amount of traffic data. In our approach, we consider P source cities denoted by $\mathcal{G}{1:P}^{\text{source}}={\mathcal{G}{1}^{\text{source}}, \cdots, \mathcal{G}_{P}^{\text{source}}}$ and one target city denoted by $\mathcal{G}^{\text {target }}$. The size of the target city is much smaller than that of the source cities, i.e., $|v^S|>>|v^T|$.

To improve prediction performance in the target city using only a small amount of data, we propose a few-shot learning method on spatio-temporal graphs. Our method leverages meta-knowledge learned from the traffic data in the source cities to enhance prediction accuracy in the target city. Specifically, we use $\mathcal{G}_{1: P}^{\text {source }}$ to train our model and extract meta-knowledge. Then, we apply this meta-knowledge to $\mathcal{G}^{\text {target}}$ to improve prediction accuracy using only a small amount of data in the target city. 

\section{METHODOLOGY}

In this section, we provide a detailed description of our STDA-Meta framework. As shown in Figure 1, STDA-Meta consists of two modules that work together for spatio-temporal knowledge transfer learning. The first module is the Spatio-Temporal Domain Adaptation (STDA) module, which includes a Spatio-Temporal Embedding (ST-E) sub-module and an Adversarial Adaptation sub-module. This module learns transferable spatio-temporal features in an adversarial way using the spatio-temporal domain loss $\mathcal{L}{st}$. The second module is the inference module, which includes the ST-E sub-module whose parameters are learned by the STDA module. The output layer of the inference module employs predictive loss $\mathcal{L}{p}$. Finally, a mixed loss function is designed to fuse the two modules. In the following subsections, we will provide a detailed description of each sub-module.

\subsection{Spatio-Temporal Embedding Module}
As shown in Figure 1, STDA is composed of spatio-temporal embedding(ST-E)  which can be divided into two temporal feature extractor(TF) and a spatial feature extractor(SF). It is formed as a “sandwich“ structure. The spatio-temporal embedding (ST-E) serves as the foundational element of the STDA framework. It captures both the spatial and temporal aspects of the input data, facilitating a comprehensive understanding of the underlying dynamics.The temporal feature extractor (TF) is a key part of the ST-E module. It focuses on extracting temporal patterns and trends from the input data, allowing for a thorough analysis of temporal evolution.

\textbf{Temporal Feature Extractor}. Although widespread for time series analysis, RNN-based models for traffic prediction still suffer from time-consuming iterations, complex gate mechanisms, and slow response to dynamic changes. On the contrary, Gated Recurrent Unit (GRU) \cite{b27} is a variant of traditional RNN. Like LSTM but in simpler structure, it can effectively capture semantic correlation between long sequences and alleviate the phenomenon of gradient disappearing or explosion. 
Take a node ${v}_{i}$ as an example, the node-level temporal meta knowledge ${z}^{tp}_{i}$ is expressed as the final state of ${h}_{i,t}$ as \ref{eq_gru}. 
Here, $d$ and $d^{\prime}$ represent the feature dimension. $x_{i, t} \in \mathbb{R}^{d}$ is the input vector of node ${v}_{i}$ at time t, and ${h}_{i,t-1}$ is the hidden state at time t-1. $W_{ir}, W_{iz}, W_{in} \in \mathbb{R}^{d \times d^{\prime}}$ and $W_{h r},W_{h z},W_{h n} \in \mathbb{R}^{d \times d^{\prime}}$ 
are weight matrices. Notations ${r}_{t},{z}_{t},{n}_{t}$ are the reset, update, and new gates respectively. The symbol $\sigma$ is the sigmoid function, and operation $*$ is the Hadamard product. Thus, we obtain a temporal meta-knowledge of a city through calculations in the Gate Recurrent Unit, denoted as $\mathbf{Z}^{t p}=\left(z_{1}^{t p}, z_{2}^{t p}, \cdots, z_{N}^{t p}\right) \in \mathbb{R}^{N \times d^{\prime}}$.

\begin{figure*}
  \includegraphics[width=\textwidth]{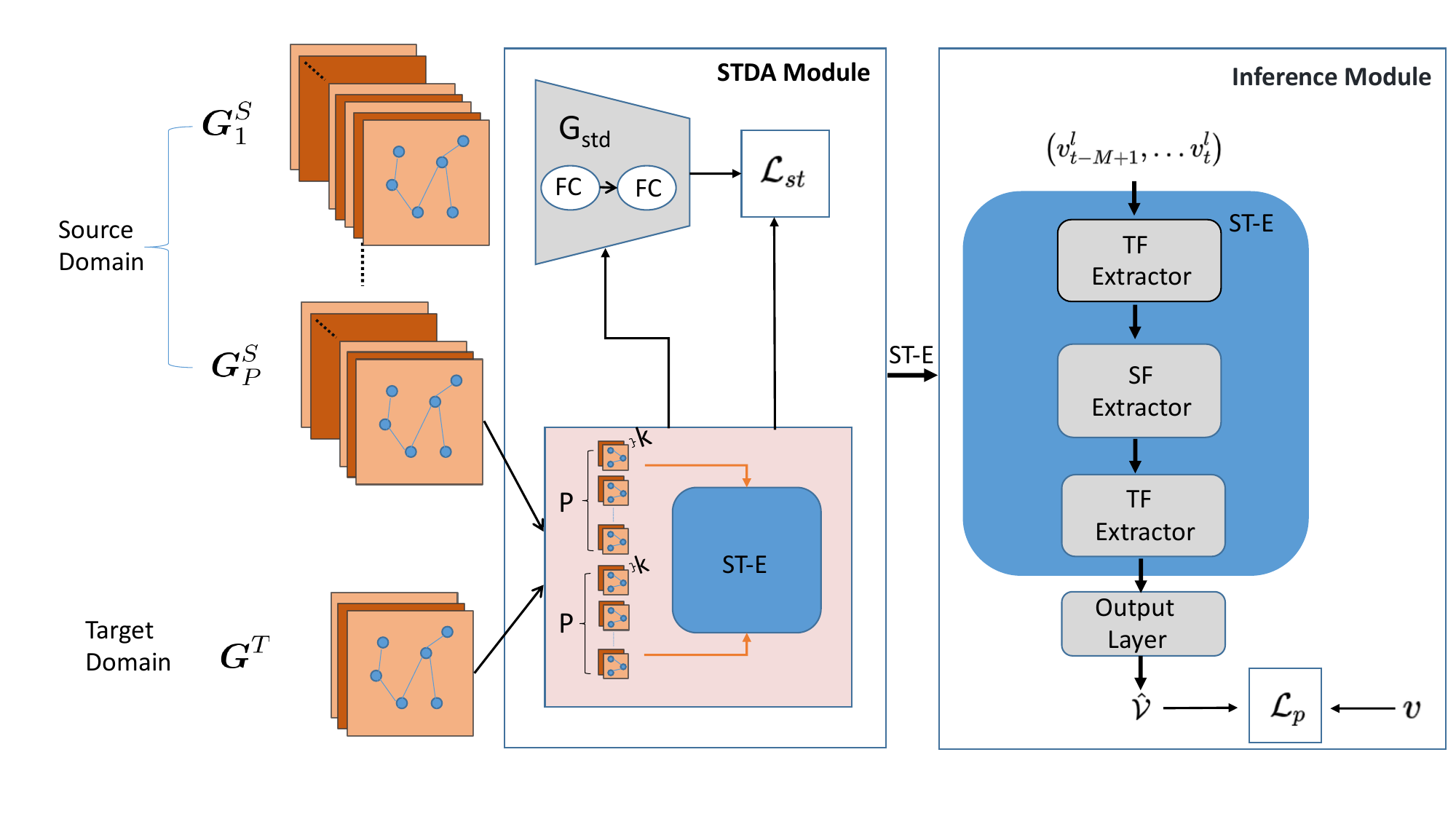}
  \caption{The Proposed STDA-Meta Framework that consists of STDA Module and Inference Module.}
\end{figure*}

\begin{equation} \label{eq_gru}
\begin{split}
r_{t} &=\sigma\left(W_{ir} x_{i,t}+b_{i r}+W_{h r} h_{i,t-1}+b_{h r}\right) \\
z_{t} &=\sigma\left(W_{iz} x_{i,t}+b_{i z}+W_{h z} h_{i,t-1}+b_{h z}\right) \\
n_{t} &=\tanh \left(W_{in} x_{i,t}+b_{i n}+r_{t} *\left(W_{h n} h_{i,t-1}+b_{h n}\right)\right) \\
  h_{i,t} &=\left(1-z_{t}\right) * n_{t}+z_{t} * h_{i,(t-1)}
\end{split}
\end{equation}

\textbf{Spatial Feature Extractor}. To capture the spatio-temporal domain-invariant correlations between different cities, we utilize the Graph Attention Network (GAT) \cite{b24} to extract the spatial correlations of spatio-temporal graphs. 
Traditional methods explicitly calculate the spatial and temporal distribution of two cities at each time step, and lead to expensive computation costs. Our method employing GAT can propose weighted summation of adjacent node features using an attention mechanism.

Compared to the Graph Convolutional Network (GCN) \cite{b25,b26}, the core difference of GAT is how to collect and aggregate the feature representation of neighbor nodes with distance 1. GAT replaces the fixed standardized operations in GCN with an attention mechanism. Essentially, GAT simply replaces the normalized function of the original GCN with a neighbor node feature aggregation function using attention weights.

In our approach, we first apply a linear transformation to each node feature, and then calculate the attention score ${e}_{ij}$ between adjacent nodes as formula \ref{eq_attention}. 
Here, $W \in \mathbb{R}^{d \times O}$ is the weight matrix and ${N}_{i}$ is a set of neighbor nodes of node ${v}_{i}$. The attention mechanism makes $\mathbb{R}^{O} \times \mathbb{R}^{O} \rightarrow \mathbb{R}$. 
The attention score is normalized then across all choices of j using the softmax fuction as formula \ref{eq_softmax}.

\begin{equation}\label{eq_attention}
e_{i j}=\text { attention }\left(W z_{i}^{t p}, W z_{j}^{t p}\right), j \in \mathcal{N}_{i} 
\end{equation}

\begin{equation}\label{eq_softmax}
\alpha_{i j}=\operatorname{softmax}_{j}\left(e_{i j}\right)=\frac{\exp \left(e_{i j}\right)}{\sum_{k \in \mathcal{N}_{i}} \exp \left(e_{i k}\right)}
\end{equation}

In order to obtain more abundant representation, we execute the attention mechanism for $K$ times independently, and employ averaging to achieve the spatial meta knowledge of node ${v}_{i}$ as formula \ref{eq_zsp}.
Thus, we obtain a spatial meta knowledge of a city, denoted as $\mathbf{Z}^{s p}=\left(z_{1}^{s p}, z_{2}^{s p}, \cdots, z_{N}^{s p}\right) \in \mathbb{R}^{N \times d^{\prime}}$. 

\begin{equation}\label{eq_zsp}
z_{i}^{s p}=\sigma\left(\frac{1}{K} \sum_{k=1}^{K} \sum_{j \in \mathcal{N}_{i}} \alpha_{i j} W^{k} z_{j}^{t p}\right)
\end{equation}

To integrate spatio-temporal features, we take $\mathbf{Z}^{s p}$ as input to temporal feature extractor and we derive the spatio-temporal knowledge of one city $\mathbf{Z}^{st p}=\left(z_{1}^{stp}, z_{2}^{stp}, \cdots, z_{N}^{stp}\right) \in \mathbb{R}^{N \times d^{\prime}}$.

\subsection{Spatio-Temporal Feature Discriminator Module}
Spatio-Temporal Feature Discriminator(${G}_{st}$) module aims to distinguish the spatio-temporal features from the source and target cities at each time step. Through adversarial classification, spatio-temporal transferable features can be captured in an efficient manner. 
   
We fetch random P days historical data from P source cities where each day contains k time intervals before the current time interval of the source cities. 
The input urban flow is $\mathbb{X}^{S}=\left\{\mathcal{X}^{(S, 1)} ; \mathcal{X}^{(S, 2)} ; \ldots ; \mathcal{X}^{(S, P)}\right\}$. Next, each crowd flow  $\mathcal{X}^{(S, i)} $ of source city at day i is fed into spatio-temporal embedding(ST-E) module. We can obtain the initial spatio-temporal feature set of the source city $Z^{S}=\left\{z_{1}^{S}, \ldots, z_{P-1}^{S}, z_{P}^{S}\right\} \in  \mathbb{R}^{P \times N \times d^{\prime}}$. Similarly, we can obtain the spatio-temporal feature set of the target city, i.e., $Z^{T}=\left\{z_{1}^{T}, \ldots, z_{P-1}^{T}, z_{P}^{T}\right\} \in  \mathbb{R}^{P \times N \times d^{\prime}}$ in the same way.
    
Based on the obtain $Z^{S}$ and $Z^{T}$, we further label these $P \times k$ samples(i.e. spatio-temporal features) as 1 or 0, where 1 is assigned to source city data as real data while 0 is assigned to target city data as fake data. Then, as depicted in Figure 1, we introduce a spatio-temporal a spatial feature discriminatory ${G}_{st}$ via fully connected(FC) layers. 
Specifcally, we mix the label $Z^{S}$ and $Z^{T}$ as the input of ${G}_{st}$. ${G}_{st}$ then distinguishes the features from the source or target city. Its gradient training function is defined as formula \ref{eq_Wd}.
Here, ${W}_{d}$ denotes the parameters of ${G}_{st}$. $z_{(i)}^{s}$ and $z_{(i)}^{t}$ represent spatio-temporal feature of source and target cities, respectively.

\begin{equation}\label{eq_Wd}
\nabla W_{d} \frac{1}{(P) \times k}\left[\log G_{s d}\left(z_{(i)}^{s}\right)+\log \left(1-G_{s d}\left({STE}\left(z_{(i)}^{t}\right)\right)\right)\right]
\end{equation}	

In terms of ${G}_{st}$ training, we add the gradient in \ref{eq_Wd} to get larger gradient for better classification performance. In contrast, ST-E aims to mix the spatio-temporal features of two cities as close as possible, so it substracts the gradient for training as formula \ref{eq_We}.
Here, ${W}_{e}$ denotes the parameters of ST-E.

\begin{equation}\label{eq_We}
    \nabla W_{e} \frac{1}{(P) \times k} \log \left(1-G_{s d}\left({STE}\left(z_{(i)}^{t}\right)\right)\right)
    \end{equation}

Through the adversarial objective, the discriminator ${G}_{st}$ is optimized to classify input spatio-temporl features into different domains, while the feature extractor ST-E is optimized in the opposite direction. Then, we define the spatio-temporal domain loss function as formula \ref{eq_Lst}.
BCELoss is widely used in classification task. The smaller the $\mathcal{L}_{st}$ the closer spatio-temporal distributions of source and target cities.
    
\begin{equation}\label{eq_Lst}
    \mathcal{L}_{st}=\operatorname{BCELoss}\left(1, G_{st}\left(STE\left(z_{(i)}^{t}\right)\right)\right)
    \end{equation}

\subsection{Inference Module}
We design the inference module on traffic prediction tasks. In that module, the input is data-scare cities time-series $v_{t-M+1}^{l}, \ldots v_{t}^{l}$, and the output one step prediction $\hat{v}$. 
As Figure. 1, in inference module, the parameters of  ST-E sub-module come from STDA module, and the output layer adopts the loss $\mathcal{L}_{p}$ defined as \ref{eq_Lp}.

\textbf{Prediction Loss} $\mathcal{L}_{p}$. Here, the Root Mean Squared Error(RMSE) is adopted as the prediction loss function. It has been widely used in various spatio-temporal prediction studies.
Here, $v$ is the ground truth and $\hat{v}$ is the predicted result.

\begin{equation}\label{eq_Lp}
\mathcal{L}_{p}=\operatorname{RMSE}(|v-\hat{v}|)
\end{equation}

\textbf{Overall Loss} $\mathcal{L}_{overall}$. 
Considering the spatio-temporal domain loss $\mathcal{L}_{st}$ and prediction loss ${L}_{p}$, we define a harmonic average as \ref{eq_overall}.
Here $\lambda$ controls the weight of spatio-temporal adaptation ability.

\begin{equation}\label{eq_overall}
\mathcal{L}_{\text {overall }}=\lambda\mathcal{L}_{st}+\mathcal{L}_{p}
\end{equation}

\subsection{STDA-Meta Learning Process}
To optimize the adaptation process in few-shot learning and enhance the adaptive ability of the model, STDA-Meta adopts a learning process that follows the Model-Agnostic Meta-Learning (MAML) based episode learning process. Specifically, STDA-Meta trains the spatio-temporal graph learning model in two stages: the base-model stage and the adaptation stage. In the base-model stage, STDA-Meta imitates the adaptation process to fine-tune the adaptation process in few-shot learning and optimize the adaptive ability. Different spatio-temporal group features are sampled into large datasets to form a batch task for the MAML-based episode, denoted as $\mathcal{T}_{S T}$. Each batch task $\mathcal{T}_{i} \in \mathcal{T}_{S T}$ includes ${K}_{S}$ support sets ${S}_{i}$ and ${K}_{Q}$ query sets ${Q}_{i}$.

In the adaptation stage, STDA-Meta updates the training parameters in the target domain through several gradient descent algorithms. Specifically, STDA-Meta first samples data from source datasets to generate batches of task sets $\mathcal{T}_{S T}$, where each task $\mathcal{T}_{i} \in \mathcal{T}_{S T}$ belongs to one single city and is divided into a support set $\mathcal{S}{\mathcal{T}_{i}}$, a query set $Q{\mathcal{T}_{i}}$, and $S{\mathcal{T}_{i}} \cap Q{\mathcal{T}_{i}}=\emptyset$. When learning a task $\mathcal{T}{i}$, STDA-Meta considers a joint loss function that combines the prediction error loss $\mathcal{L}{\mathcal{T}_{i}}$ as formula \ref{eq_LT}.
Here, $\mathcal{L}{\mathcal{T}_{i}}$ is the root mean square error between multi-step prediction and the ground truth of the support set $S_{\mathcal{T}_{i}}$.

\begin{equation}\label{eq_LT}
\mathcal{L}{\mathcal{T}_{i}}=\frac{1}{\left|\mathcal{S}{\mathcal{T}{i}}\right|} \sum_{\left(x_{j}, y_{j}\right) \in \mathcal{S}{\mathcal{T}{i}}}\left(f_{\theta}\left(x_{j}\right)-y_{j}\right)^{2},
\end{equation}

To achieve the optimal model parameter $\theta^{*}$, STDA-Meta adopts the MAML optimization algorithm. Specifically, the base-model stage of STDA-Meta involves iteratively fine-tuning the model on a small set of examples sampled from the support set of each task. Then, the adapted model is tested on the corresponding query set, and the parameters are updated by back-propagating the prediction error through the adapted model. 
This process is repeated for each task, and the model with the lowest prediction error on the query set is selected as the final model. 
Overall, the MAML-based episode learning process enables STDA-Meta to optimize the adaptation process in few-shot learning and enhance the adaptive ability of the model.

\section{METHODOLOGY}


\section{Experiments}
\textbf{Datasets.} We utilized four widely used open-source traffic datasets from different cities, including METR-LA, PEMS-BAY, Didi-Chengdu, and Didi-Shenzhen \cite{b28,b29}. 
These datasets have been widely adopted in related studies and contain valuable information for traffic speed prediction. 
Table I summarizes the static information of the datasets, including the number of nodes, edges, interval, and time span.

Nodes here is the number of traffic sensors or detectors in the road network. Edges represent the links between nodes in the network. Interval refers to the time interval between two consecutive data points in the dataset. Time span represents the total time period covered by the dataset. Please see Table I for the detailed information of the datasets.
Interval represents the time interval between two consecutive data points in the dataset. Time span represents the total time period covered by the dataset. 

 \begin{table}
 \caption{Static Information of Datasets}
\begin{tabular}{lllll}
\hline
Citys      & Didi-Shenzhen & Didi-Chengdu& PEMS-BAY &METR-LA  \\ \hline
Nodes      & 627           & 524         & 325      & 207     \\
Edges      & 4845          & 1120        & 2694     & 1722    \\
Interval   & 5 min          & 5 min       & 5 min     & 5 min   \\
Time span  & 17280         & 17280       & 52116    & 34272   \\ \hline
\end{tabular}
\end{table} 

\textbf{Data Processing.} During the data processing stage, the datasets are preprocessed by normalizing them using the Z-Score method. To evaluate the ability of STDA-Meta for spatio-temporal knowledge transfer, the datasets are partitioned into three subsets, namely training, validation, and testing sets. As an example, we consider REMS-BAY as the target city, and select only three days of data (which is a very small subset compared to the complete data of other cities) as the target data for that city. The remaining data from other cities, namely METR-LA, Didi-Chengdu, and Didi-Shenzhen, are used as the source cities for meta-training. The test set for REMS-BAY consists of data from the remaining days.

\textbf{Experimental Settings.}
All experiments are compiled and tested on a server (CPU:Intel(R) Xeon(R) Platinum 8163 CPU @ 2.50GHz, GPU:NVIDIA Corporation TU104GL [Tesla T4]). 
In order to more fully test the performance of our framework, we predict the traffic speed in the next 6 time steps with 12 historical time steps. 
Some important parameters are set as follows: task learning rate $\alpha = 0.01$, meta-training rate $\beta = 0.001$, task batch number $\|\mathcal{T}\|=5$ and sum scale factor of $\mathcal{L}_{overall}$ $\lambda = 1.5$. 
Metrics Root Mean Squared Error (RMSE) and Mean Absolute Error (MAE), widely used in traffic prediction, are evaluated here  as formula \ref{eq_metrics}. Here, $y_{i}$ represents the actual value, $y_{p}$ represents the predicted value, and n represents the number of observations.

\begin{equation}\label{eq_metrics}
\begin{split}
\text { MAE }=\frac{\left|\left(y_{i}-y_{p}\right)\right|}{n}\\
\text { RMSE }=\sqrt{\frac{\sum\left(y_{i}-y_{p}\right)^{2}}
{n}}
\end{split}
\end{equation}

\textbf{Baselines}
We compare our framework STDA-Meta with the following two types of baselines. 

1). Classic spatio-temporal graph learning methods. We choose LSGCN \cite{b30} and STGCN \cite{b31} as they are remarkable models for crowd flow prediction. Here, we only use target city data for model training.

2). Transfer learning methods. These methods first pre-train the deep models on the source city and then fine-tune models on the target city. We use LSGCN \cite{b30} and STGCN \cite{b31} as the base-model and we call them LSGCN-FT and STGCN-FT. AdaRNN \cite{b38} is a state-of-the-art transfer learning framework for non-stationary time series.


\textbf{Performance Comparison}
In Table 2, each row represents a different method categorized into three types: Spatio-temporal graph learning methods, Transfer Learning methods, and variants of our proposed approach (STDA-Meta). 
We compared these methods across two distinct experimental settings. 
One setting involved using METR-LA, Didi Chengdu, and Didi Shenzhen as source cities, with PEMS-BAY as the target city. 
The other setting utilized PEMS-BAY, Didi Chengdu, and Didi Shenzhen as source cities, with METR-LA as the target city.

The performance of spatio-temporal graph learning methods was relatively poor due to the scarcity of samples in the target city, which constrained their ability to learn effective models. 
These methods typically demand a substantial volume of data for model training, and when the target city has limited samples, their performance is notably affected.

Although the fine-tuned baseline models exhibited improved performance compared to non-transfer models, they still lagged behind transfer learning methods. Fine-tuning can moderately adjust the model to suit the target city's data, but its efficiency is hampered by the availability and diversity of data in the target city.

Although the AdaRNN method performs well in transfer learning, particularly for forecasting time series with limited samples, it exhibits limitations in capturing spatial information within the complex scenario of overall traffic prediction. This constraint hinders its performance, making it slightly inferior to the performance of STDA-Meta, especially in scenarios involving comprehensive traffic prediction across the entire road network. 
AdaRNN's inability to thoroughly capture spatial features diminishes its effectiveness in traffic prediction tasks that encompass diverse geographical regions. In contrast, STDA-Meta, leveraging its superior meta-learning strategy, efficiently utilizes information from source cities and comprehensively understands and exploits the intricate traffic patterns of the target city. Thus, STDA-Meta maintains superior performance in the context of overall traffic prediction, presenting a comparative advantage over the AdaRNN method.

It's noteworthy that STDA-Meta consistently outperformed all baseline models in both experimental settings. It can be attributed to STDA-Meta's superior ability to harness information from the source cities, facilitating efficient learning for the target city. 
Here, our model achieved a 7\% enhancement in predictive performance compared to the baseline models, underscoring its superiority.

STDA-Meta, utilizing a meta-learning approach, showcases improved predictive performance by adapting to the data of the target city, emphasizing notable enhancements in performance.

\textbf{Ablation Study}
Through the results above, the RMSE and MAE show a slight increase after removing either Domain Adaptation (DA) or the MAML (Model-Agnostic Meta-Learning) method, but both variants still outperform the baseline models.
We further elaborate on the performance impact of DA and MAML, and the results are presented in Table 2. Specifically, we discuss the performance of STDA-Meta w/o DA and STDA-Meta w/o Meta.

1) The performance impact on DA.
The decrease in performance when removing Domain Adaptation (DA) can be attributed to its role in facilitating effective domain transfer. DA enables the model to adapt to the target domain's characteristics and data distribution. 
The absence of DA makes the model hard to adapt to the target domain, and a slight increases appear in both RMSE and MAE.

2) The performance impact on MAML.
The decrease in performance when removing the MAML (Model-Agnostic Meta-Learning) can be explained here. MAML enables the model to quickly adapt to new tasks or domains, and enhances the model's generalization capabilities. It is particularly crucial for few-shot learning scenarios. Removing MAML eliminates the mechanism for rapid adaptation, and declines in performance appear. 

In summary, the experimental results highlight the crucial roles of Domain Adaptation and Model-Agnostic Meta-Learning in performance. Removing either component would weaken the model's capabilities, but the variants  still outperform baselines. It proves the effectiveness and necessity of these key components.

\textbf{Hyperparameter Analysis}.
We conducted a series of experiments to determine the optimum of hyperparameter $\lambda$ in our model. By adjusting $\lambda$ from 0.5 to 2.0 and repeating the experiments, we obtained the results presented in Figure 2. From the results in this figure, the weight ratio of the two loss functions in $\mathcal{L}_{overall}$ is an important factor to consider, and the choice of $\lambda$ significantly impacts the model's prediction performance. Specifically, we observed that increasing the value of $\lambda$ often improves predictive results. It highlights the importance of the domain adaptation (DA) module during optimizing the model. It's worth noting that our findings indicate that incorporating data augmentation (DA) techniques into machine learning models can significantly improve performance in data-scarce scenarios.
Our approach of adjusting the hyperparameter $\lambda$ to optimize the balance between the two loss functions is simple yet effective to achieve better results.

Furthermore, we also observed that the choice of $\lambda$ is related to the time scale. When the time scale is relatively long, an appropriate $\lambda$ value is crucial. The predictive accuracy tends to increase initially with the increase in $\lambda$, followed by a decrease. It indicates such a hyperparameter tuning in long time series prediction is significant.

In summary, key hyperparameter is essential to achieve optimal performance in our method.


\section{Discussion and Future Work}
In this paper, we propose a novel Meta-Learning framework, namely STDA-Meta, for few-shot traffic prediction. The STDA module, which is based on city-level meta knowledge, enhances the effectiveness of spatio-temporal representation on multiple datasets. STDA-Meta integrates MAML-based episode learning process to learn easily adaptable model parameters through gradient descent. Extensive experimental results on traffic speed data demonstrate the superiority of STDA-Meta over other baseline methods.

In fact, the proposed STDA-Meta framework is not only limited in traffic prediction but can be applied to other few-shot scenarios involving spatio-temporal graph learning, such as residential electric load forecasting and taxi demand prediction in different cities. 
In future, we aim to further exploit and extend STDA-Meta in other few-shot learning tasks to enhance predictive performance.

\begin{table*}[htb!]
\center

\caption{Performance of STDA-Meta model under PEMS-BAY and METR-LA data sets, where the best results are underlined.}
\resizebox{\textwidth}{20mm} {
\begin{tabular}{|cc|cccccc|cccccc|}

\hline
\multicolumn{2}{|c|}{}                                                        & \multicolumn{6}{c|}{\textbf{METR-LA}}                                                                         & \multicolumn{6}{c|}{\textbf{PEMS-BAY}}                                                                          \\ \cline{3-14} 
\multicolumn{2}{|c|}{\textbf{Methods}}                                        & \multicolumn{3}{c|}{\textbf{MAE}}                                & \multicolumn{3}{c|}{\textbf{RMSE}}          & \multicolumn{3}{c|}{\textbf{MAE}}                                & \multicolumn{3}{c|}{\textbf{RMSE}}          \\ \cline{3-14} 
\multicolumn{2}{|c|}{}                                                        & \textbf{5min} & \textbf{15min} & \multicolumn{1}{c|}{\textbf{30min}} & \textbf{5min} & \textbf{15min} & \textbf{30min} & \textbf{5min} & \textbf{15min} & \multicolumn{1}{c|}{\textbf{30min}} & \textbf{5min} & \textbf{15min} & \textbf{30min} \\ \hline
\multicolumn{1}{|c|}{\textbf{Spatio-temporal graph learning}}     & \textbf{LSGCN}             & 1.703 & 2.031 &\multicolumn{1}{c|}{2.224}        & 3.042 & 3.153 & 3.733      & 1.997 & 2.420 &\multicolumn{1}{c|}{2.594}    & 2.970 & 3.177 & 3.351         \\ \cline{2-14} 
\multicolumn{1}{|c|}{\textbf{}}                  & \textbf{STGCN}             & 1.523 & 1.816 &\multicolumn{1}{c|}{2.003}        & 3.070 & 3.210 & 3.325      & 2.111 & 2.263 &\multicolumn{1}{c|}{2.628}    & 2.804 & 3.026 & 3.373               \\ \hline
\multicolumn{1}{|c|}{\textbf{}}                  & \textbf{LSGCN-FT}          & 1.270 & 1.396 &\multicolumn{1}{c|}{1.739}        & 2.821 & 2.982 & 3.519      & 1.768 & 1.872 &\multicolumn{1}{c|}{2.011}    & 2.913 & 3.137 & 3.319               \\ \cline{2-14} 
\multicolumn{1}{|c|}{\textbf{Transfer learning}} & \textbf{STGCN-FT}          & 1.172 & 1.415 &\multicolumn{1}{c|}{1.598}        & 2.870 & 2.987 & 3.137      & 1.646 & 1.918 &\multicolumn{1}{c|}{2.252}    & 2.607 & 2.718 & 2.953               \\ \cline{2-14} 
\multicolumn{1}{|c|}{\textbf{}}                  & \textbf{AdaRNN}            & 0.954 & 1.150 &\multicolumn{1}{c|}{1.252}        & 2.377 & 2.492 & 2.856      & 1.443 & 1.676 &\multicolumn{1}{c|}{1.784}    & 2.343 & 2.590 & 2.700               \\ \hline
\multicolumn{1}{|c|}{\textbf{}}                  & \textbf{STDA-Meta w/o DA}   & 1.153 & 1.289 &\multicolumn{1}{c|}{1.634}        & 2.244 & 2.470 & 2.672      & 1.323 & 1.663 &\multicolumn{1}{c|}{1.788}    & 2.301 & 2.508 & 2.609               \\ \cline{2-14} 
\multicolumn{1}{|c|}{\textbf{ours methods}}      & \textbf{STDA-Meta w/o Meta} & 0.742 & 1.185 &\multicolumn{1}{c|}{1.609}        & 2.150 & 2.440 & 2.839      & 1.221 & 1.439 &\multicolumn{1}{c|}{1.800}    & 1.995 & \underline{2.138} & 2.660               \\ \cline{2-14} 
\multicolumn{1}{|c|}{\textbf{}}                  & \textbf{STDA-Meta}         & \underline{0.524} & \underline{0.816} &\multicolumn{1}{c|}{ \underline{0.998}}        & \underline{2.011} & \underline{2.292} & \underline{2.399}      & \underline{1.101} & \underline{1.398} &\multicolumn{1}{c|}{\underline{1.638}}    & \underline{2.163} & 2.309 & \underline{2.452}               \\ \hline
\end{tabular}
}
\end{table*}

\begin{figure*}[htp!]
  \includegraphics[width=\textwidth]{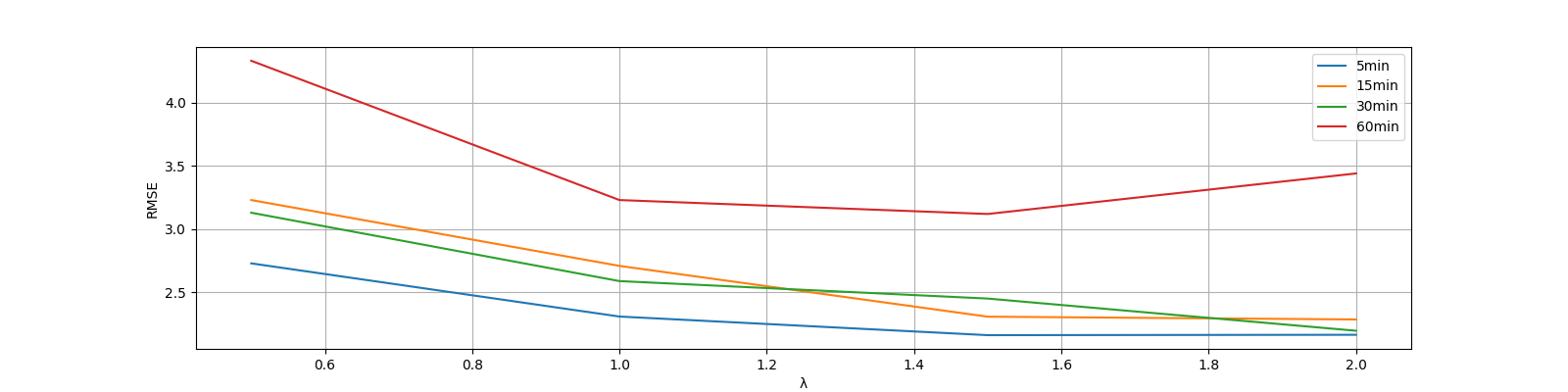}
  \caption{Hyperparameter study on PEMS-BAY dataset: sum scale factor $\lambda$}
\end{figure*}


\begin{thebibliography}{00}
\bibitem{b1} Yang, Haiqiang, et al. "Region-Level Traffic Prediction Based on Temporal Multi-Spatial Dependence Graph Convolutional Network from GPS Data." Remote Sensing 14.2 (2022): 303.
\bibitem{b2} Han, Ying, et al. "Multi-step network traffic prediction using echo state network with a selective error compensation strategy." Transactions of the Institute of Measurement and Control 44.8 (2022): 1656-1668.
\bibitem{b3} Kumar, P. M., Basheer, S., Rawal, B. S., Afghah, F., Babu, G. C., Arunmozhi, M. (2022). Traffic scheduling, network slicing and virtualization based on deep reinforcement learning. Computers and Electrical Engineering, 100, 107987.
\bibitem{b4} Kumar, Neetesh, et al. "Deep Reinforcement Learning-Based Traffic Light Scheduling Framework for SDN-Enabled Smart Transportation System." IEEE Transactions on Intelligent Transportation Systems 23.3 (2021): 2411-2421.
\bibitem{b5} Mohanty, Sudatta, Alexey Pozdnukhov, and Michael Cassidy. "Region-wide congestion prediction and control using deep learning." Transportation Research Part C: Emerging Technologies 116 (2020): 102624.
\bibitem{b6} Pawan Lingras, Satish C. Sharma, Phil Osborne and Iftekhar Kalyar: ”Traffic Volume Time-Series Analysis According to the Type of Road Use”, Computer-Aided Civil and Infrastructure Engineering, 2000, 15, (5), pp. 365-373
\bibitem{b7} M. M. Hamed, H. R. Al-Masaeid, and Z. M. B. Said, “Short-term prediction of traffic volume in urban arterials,” Journal of Transportation Engineering, vol. 121, no. 3, pp. 249–254, 1995.
\bibitem{b8} Weilong Ding, Zhuofeng Zhao, ”DS-Harmonizer: A Harmonization Service on Spatiotemporal Data Stream in Edge Computing Environment”, Wireless Communications and Mobile Computing, vol. 2018, Article ID 9354273, 12 pages, 2018. https://doi.org/10.1155/2018/9354273
\bibitem{b9} Ian J. Goodfellow, Jean PougetAbadie, Mehdi Mirza, Bing Xu, David Warde-Farley,
Sherjil Ozair, Aaron C. Courville, and Yoshua Bengio. enerative adversarial nets. In NeurIPS, 2014.
\bibitem{b10}Wang, Haoxiang, et al. "Global Convergence of MAML and Theory-Inspired Neural Architecture Search for Few-Shot Learning." Proceedings of the IEEE/CVF Conference on Computer Vision and Pattern Recognition. 2022.
\bibitem{b11}E. Zivot and J. Wang, “Vector autoregressive models for multivariate time series,” Modeling Financial Time Series with S-Plus R , pp. 385–429, 2006.
\bibitem{b12}Y. Lv, Y. Duan, W. Kang, Z. Li and F. Y. Wang: ”Traffic Flow Prediction With Big Data: A Deep Learning Approach”, IEEE Transactions on Intelligent Transportation Systems, 2015, 16, (2), pp. 865-873
\bibitem{b13}Yonghong Luo, Xiangrui Cai, Ying Zhang, Jun Xu, et al. Multivariate time series imputation with generative adversarial networks. In Advances in Neural Information Processing Systems, pages 1596–1607, 2018.
\bibitem{b14}Syama Sundar Rangapuram, Matthias W Seeger, Jan Gasthaus, Lorenzo Stella, Yuyang Wang, and Tim Januschowski. Deep state space models for time series forecasting. In Advances in neural information processing systems, pages 7785–7794, 2018.
\bibitem{b15}Junbo Zhang, Yu Zheng, Dekang Qi, Ruiyuan Li, and Xiuwen Yi. Dnn-based prediction model for spatio-temporal data. In Proceedings of the 24th ACM SIGSPATIAL International Conference on Advances in Geographic Information Systems, pages 1–4, 2016.
\bibitem{b16}Michaël Defferrard, Xavier Bresson, and Pierre Vandergheynst. Convolutional neural networks on graphs with fast localized spectral filtering. In Advances in neural information processing systems, pages 3844–3852, 2016.
\bibitem{b17}Zhao, Ling, et al. "T-gcn: A temporal graph convolutional network for traffic prediction." IEEE Transactions on Intelligent Transportation Systems 21.9 (2019): 3848-3858.
\bibitem{b18}Han, Xu, and Shicai Gong. "LST-GCN: Long Short-Term Memory Embedded Graph Convolution Network for Traffic Flow Forecasting." Electronics 11.14 (2022): 2230.
\bibitem{b19}Chen, Zhi, Fan Yang, and Wenbing Tao. "Detarnet: Decoupling translation and rotation by siamese network for point cloud registration." Proceedings of the AAAI Conference on Artificial Intelligence. Vol. 36. No. 1. 2022.
\bibitem{b20}Zheng, Wenfeng, et al. "A deep fusion matching network semantic reasoning model." Applied Sciences 12.7 (2022): 3416.
\bibitem{b21}Zhou, Fan, et al. "Meta-gnn: On few-shot node classification in graph meta-learning." Proceedings of the 28th ACM International Conference on Information and Knowledge Management. 2019.
\bibitem{b22}Liu, Zemin, et al. "Relative and absolute location embedding for few-shot node classification on graph." Proceedings of the AAAI conference on artificial intelligence. Vol. 35. No. 5. 2021.
\bibitem{b23}Ding, Kaize, et al. "Few-shot network anomaly detection via cross-network meta-learning." Proceedings of the Web Conference 2021. 2021.
\bibitem{b24}HongHao Gao, Junsheng Xiao, Yuyu Yin, Tong Liu, Jiangang Shi. A Mutually Supervised Graph Attention Network for Few-shot Segmentation: The Perspective of Fully Utilizing Limited Samples. IEEE Transactions on Neural Networks and Learning Systems(TNNLS), 2022, DOI:10.1109/TNNLS.2022.3155486.
\bibitem{b25}Velickovic, Petar, Guillem Cucurull, Arantxa Casanova, Adriana Romero, Pietro Lio, and Yoshua Bengio. "Graph attention networks." stat 1050 (2017): 20.
\bibitem{b26}Han, Shi-Yuan, Qiang Zhao, Qi-Wei Sun, Jin Zhou, and Yue-Hui Chen. "EnGS-DGR: Traffic Flow Forecasting with Indefinite Forecasting Interval by Ensemble GCN, Seq2Seq, and Dynamic Graph Reconfiguration." Applied Sciences 12, no. 6 (2022): 2890.
\bibitem{b27}Zhang, Kun, et al. "A GRU-based ensemble learning method for time-variant uncertain structural response analysis." Computer Methods in Applied Mechanics and Engineering 391 (2022): 114516.
\bibitem{b28}Yu, H., Xu, X., Zhong, T., Zhou, F. (2022). Fine-grained urban flow inference via normalizing flows (student abstract). In AAAI.
\bibitem{b29}Lu, Bin, Xiaoying Gan, Weinan Zhang, Huaxiu Yao, Luoyi Fu, and Xinbing Wang. "Spatio-Temporal Graph Few-Shot Learning with Cross-City Knowledge Transfer." arXiv preprint arXiv:2205.13947 (2022).
\bibitem{b30}Huang, Rongzhou, et al. "LSGCN: Long Short-Term Traffic Prediction with Graph Convolutional Networks." IJCAI. 2020.
\bibitem{b31}Yu, Bing, Haoteng Yin, and Zhanxing Zhu. "Spatio-temporal graph convolutional networks: A deep learning framework for traffic forecasting." arXiv preprint arXiv:1709.04875 (2017).
\bibitem{b32}J. Yuan, Y. Zheng, X. Xie, and G. Sun, “Driving with knowledge from the physical world,” in Proc. 17th ACM SIGKDD Int. Conf. Knowl. Discovery Data Mining (KDD), 2011, pp. 316–324.
\bibitem{b33}Jie Zhou, Weilong Ding, Zhuofeng Zhao and Han Li: ”SMART: A Service-Oriented Statistical Analysis Framework on Spatio-Temporal Big Data (Short Paper)”. Proc. 15th International Conference on Collaborative Computing: Networking, Applications and Worksharing (CollaborateCom 2019). Springer International Publishing. London, Great Britain, 2019, pp. 91-100. (EI, CCF C, Accession number: 20200107954781).
\bibitem{b34}Fang, Ziquan, Dongen Wu, and Lu Pan. "When Transfer Learning Meets Cross-City Urban Flow Prediction: Spatio-Temporal Adaptation Matters." IJCAI’22 (2022): 2030-2036.
\bibitem{b35}Lu, Bin, et al. "Spatio-Temporal Graph Few-Shot Learning with Cross-City Knowledge Transfer." arXiv preprint arXiv:2205.13947 (2022).
\bibitem{b36}Honghao Gao, Danqing Fang, Junsheng Xiao, Walayat Hussain, Jung Yoon Kim. CAMRL: A Joint Method of Channel Attention and Multidimension Regression Loss to 3D Object Detection for Automated Vehicle. IEEE Transactions on Intelligent Transportation Systems(T-ITS),2022, DOI:10.1109/TITS.2022.3219474
\bibitem{b37}Yao H, Zhang C, Wei Y, et al. Graph few-shot learning via knowledge transfer[C]//Proceedings of the AAAI Conference on Artificial Intelligence. 2020, 34(04): 6656-6663.
\pagebreak
\bibitem{b38}Du Y, Wang J, Feng W, et al. Adarnn: Adaptive learning and forecasting of time series[C]Proceedings of the 30th ACM international conference on information  knowledge management. 2021: 402-411.


\enlargethispage{-20mm}

\end{thebibliography}
\end{document}